\documentclass[conference]{IEEEtran}
\usepackage{cite}
\usepackage{textcomp}
\usepackage{xcolor}
\usepackage[numbers]{natbib}
\usepackage{color}

\usepackage{amsmath}
\usepackage{amsfonts}
\usepackage{amsthm}
\usepackage{bm}
\usepackage{algorithm}
\usepackage[noend]{algpseudocode}

\usepackage{multirow} 
\usepackage{booktabs}

\usepackage{subcaption}
\usepackage{caption}
\usepackage{graphicx}
\usepackage{enumitem}

\usepackage{balance}

\newcommand{\revise}[1]{{#1}}

\def\BibTeX{{\rm B\kern-.05em{\sc i\kern-.025em b}\kern-.08em
    T\kern-.1667em\lower.7ex\hbox{E}\kern-.125emX}}

\IEEEoverridecommandlockouts

\begin{document}

\title{DoubleEnsemble: A New Ensemble Method Based on Sample Reweighting and Feature Selection for Financial Data Analysis}


\author{\IEEEauthorblockN{1\textsuperscript{st} Chuheng Zhang \thanks{This work was done when Chuhang Zhang and Yuanqi Li were interns with X-Tech Co. Ltd.}}
\IEEEauthorblockA{\textit{Tsinghua University}\\
Beijing, China \\
zhangchuheng123@live.com}
\\
\IEEEauthorblockN{4\textsuperscript{th} Yifei Jin}
\IEEEauthorblockA{\textit{Tsinghua University, X-Tech Co. Ltd.}\\
Beijing, China \\
yfjin1990@gmail.com}
\and
\IEEEauthorblockN{1\textsuperscript{st} Yuanqi Li}
\IEEEauthorblockA{\textit{E-hualu}\\
Beijing, China \\
liyq@ehualu.com}
\\
\IEEEauthorblockN{5\textsuperscript{th} Pingzhong Tang}
\IEEEauthorblockA{\textit{Tsinghua University} \\
Beijing, China \\
kenshinping@gmail.com}
\and
\IEEEauthorblockN{3\textsuperscript{rd} Xi Chen}
\IEEEauthorblockA{\textit{New York University}\\
New York, USA \\
xc13@stern.nyu.edu}
\\
\IEEEauthorblockN{6\textsuperscript{th} Jian Li}
\IEEEauthorblockA{\textit{Tsinghua University}\\
Beijing, China \\
lijian83@mail.tsinghua.edu.cn}
}

\maketitle

\begin{abstract}
    Modern machine learning models (such as deep neural networks and boosting decision tree models) have become increasingly popular in financial market prediction, due to their superior capacity to extract complex non-linear patterns. However, since financial datasets have very low signal-to-noise ratio and are non-stationary, complex models are often very prone to overfitting and suffer from instability issues. Moreover, as various machine learning and data mining tools become more widely used in quantitative trading, many trading firms have been producing an increasing number of features (aka factors). Therefore, how to automatically select effective features becomes an imminent problem. To address these issues, we propose DoubleEnsemble, an ensemble framework leveraging learning trajectory based sample reweighting and shuffling based feature selection. 
    Specifically, we identify the key samples based on the training dynamics on each sample and elicit key features based on the ablation impact of each feature via shuffling.
    Our model is applicable to a wide range of base models, capable of extracting complex patterns, while mitigating the overfitting and instability issues for financial market prediction. We conduct extensive experiments, including price prediction for cryptocurrencies and stock trading, using both DNN and gradient boosting decision tree as base models. Our experiment results demonstrate that DoubleEnsemble achieves a superior performance compared with several baseline methods.
\end{abstract}


\section{Introduction}

Financial market is notoriously difficult to predict due to its competing nature. There are some common reasons that partially explain why the prediction task is extremely difficult. First, the difficulty comes from the widely known efficient market theory, 
which is a hypothesis that states that share prices reflect all information and  it is impossible to consistently outperform the overall market (see e.g., the original paper by \citeauthor{samuelson2016proof} \cite{samuelson2016proof}). Second, due to existence of a large number of ``noisy traders'' \cite{bloomfield2009noise}, and other hidden factors that impact the movement of the market (e.g., government policy changes and breaking news), the financial data is highly noisy, dynamic and volatile.

Multifactor model \cite{ross1976arbitrage} is a popular model for asset pricing and market prediction.
The model prices the asset or predicts the market movement based on multiple features (or factors), 
such as the firm size \cite{banz1981relationship}, the earnings' yield \cite{basu1983relationship}, the leverage \cite{bhandari1988debt} and the book-to-market ratio \cite{chan1991fundamentals}.
Linear model has been a standard algorithm for the multifactor model but has a great limitation in exploiting complex patterns.
Recently, non-linear machine learning models (such as gradient boosting decision trees or deep learning models) become popular due to their large model capacity \cite{ochotorena2012robust, arevalo2016high, deng2016deep, fischer2018deep, jia2019quantitative}.
However, these complex non-linear models are prone to overfitting and susceptible to noisy samples.

To provide the model with more information, 
quantitative traders or researchers often create hundreds or even thousands of features (aka factors) \cite{kakushadze2016101,de2018advances,feng2018deep,zhang2020autoalpha}. 
However, training a prediction model with all the available features may lead to poor performance.
Therefore, it is essential to select features that are not only informative but also uncorrelated with other features.
For linear models (such as linear regression), we can select features with low correlations to alleviate the multicollinearity problem (see e.g., \cite{farrar1967multicollinearity}). For highly complex non-linear models and highly noisy financial data, it is less clear how to effectively select features.

To address the aforementioned issues, we propose DoubleEnsemble, a new ensemble framework for financial market prediction. In particular, we construct sub-models in the ensemble one by one, where each sub-model 
is trained with
both the weights of samples and carefully selected features. 
A wide range of base models can be used in learning the sub-models, such as the linear regression model, boosting decision trees, and deep neural networks. 
Each time, using our \emph{learning trajectory based sample reweighting} scheme, we assign a weight 
to each sample in the original training set based on 
\revise{the loss curve of the previous sub-model and the loss value of the current ensemble (which we refer to as the learning trajectory).}
Moreover, we select features based on their contribution to the current ensemble via a \emph{shuffling technique}. 


There are three major contributions/features of our proposed DoubleEnsemble framework. 
\begin{enumerate}
    \item 
    Our method integrates sample reweighting and feature selection into a unified framework, and is named DoubleEnsemble. 
    We ensemble  diversified sub-models that are trained with not only different sample weights but also features. 
    This property greatly alleviates the overfitting problem and makes DoubleEnsemble more stable and suitable for learning from highly noisy financial data.
    \item For the sample reweighting component, we propose a new \emph{learning trajectory based sample reweighting} scheme, which fully incorporates 
    \revise{the learning trajectory}
    into the construction of sample weights. This reweighting scheme
    can effectively reduce the weights of very easy and noisy samples 
    and boosts that of the key samples that are more informative for training the model.
    \footnote{
    Easy samples are those which the algorithm can classify correctly very easily. Fitting pure noisy samples may lead to overfitting. Hence, we would like the learning algorithm
    to focus less on these and more on the remaining samples.
    See Section~\ref{sec:SR} for the details.
    }
    \item  For feature selection, traditional approaches (e.g., backward elimination and recursive feature elimination) usually attempt to remove redundant features according to their importance and retrain the whole model after removing each feature.
In practice, retraining incurs a huge computational cost. Moreover, when training with neural networks, removing a feature could completely change the distribution of inputs, which leads to extremely unstable training process.  
To address the challenge,  we propose a new \emph{shuffling based feature selection method}. Instead of removing a feature, we shuffle a feature across training samples and measure the change of the loss. The small change indicates that the feature is less relevant for the predication task. Our feature selection approach is both computationally efficient and has shown to be effective on real financial datasets with a large number of factors.
\end{enumerate}

In the experiments, we apply DoubleEnsemble to two financial markets, the cryptocurrency exchange OKEx and the securities exchange China's A-share market. 
These two markets possess different trading rules and market participants, and therefore there are different types of noise and patterns in the historical data of these two markets.
Moreover, we use DoubleEnsemble to construct prediction models to trade at different frequencies (from seconds to weeks).
Our experiments show that DoubleEnsemble achieves superior performances in both markets. Specifically, DoubleEnsemble achieves a precision of 62.87\% for predicting the direction of the cryptocurrency movement and an annualized return over 51.37\% with the Sharpe ratio 4.941 in China's A-share market.

The rest of our paper is organized as follows. We introduce the related work in Section \ref{sec:related_work}. Then, we introduce DoubleEnsemble in Section \ref{sec:method} and present the experiment results in Section \ref{sec:experiment}. At last, we summarize our work in Section \ref{sec:conclusion}.

\section{Related Work}
\label{sec:related_work}

\textbf{Ensemble Model.}
Ensemble is an effective way to enhance the model robustness. 
The key for an ensemble model is to construct \emph{good} and \emph{diverse} sub-models.
The methods to construct sub-models can be divided into two categories.
In the first category, individual but different models can be built separately, such as bagging \cite{breiman1996bagging}. 
This category is popular for financial market prediction. 
For example, \citeauthor{liang2012stock} \cite{liang2012stock} use different base models to construct different sub-models; \citeauthor{xiang2006predicting} \cite{xiang2006predicting} and \citeauthor{zhai2010hybrid} \cite{zhai2010hybrid} construct sub-models by selecting financial data from different time periods or different market environments respectively. 
The other category builds the sub-models based on the performance of those built previously, such as boosting \cite{freund1995desicion}.
The model built through this category of methods has better predictive accuracy but tends to overfit to the noise in the training data \cite{long2010random} and therefore is not currently widely used for financial market prediction. 

\textbf{Sample reweighting.}
Weighting the samples for the model training 
is shown to be effective in some computer vision applications:
\citeauthor{saxena2019data} \cite{saxena2019data} treat the weights of the samples as parameters and learn the weights via the gradient. 
\citeauthor{hu2019learning} \cite{hu2019learning} and \citeauthor{fan2018learning} \cite{fan2018learning} design a reward function for the weights and learn the weights via reinforcement learning. 
\citeauthor{ren2018learning} \cite{ren2018learning} train an additional neural network to learn the weights. 

There is a conflict between the objective of boosting and denoising when assigning weights to the samples for the model training. 
Boosting increases the weights of the hard samples.
This is similar to curriculum learning \cite{bengio2009curriculum} where the model is trained to first fit the easy samples and then the hard samples. 
In financial market prediction, this can also be interpreted as learning another new pattern when the previous patterns are exploited. 
Examples of this trend of reweighting are \cite{saxena2019data} and \cite{fan2018learning}. 
On the other hand, for constructing an ensemble robust to the outliers and noisy samples, weights of these samples should be reduced. 
For instance, \citeauthor{jiang2018mentornet} \cite{jiang2018mentornet}, \citeauthor{liu2019self} \cite{liu2019self} and \citeauthor{nguyen2019self} \cite{nguyen2019self} reduce the weights of the samples that the model does not fit well. 
However, it is hard for us to distinguish between the hard samples and the outliers or the noisy samples.
It is a challenge to reduce the weights of noisy samples while performing a boosting style of learning. 

\textbf{Feature selection.}
Conventionally, features for financial market prediction are manually selected \cite{kwon2007hybrid,luo2013integrating}. 
However, automation for feature selection is desired when the number of features increases. 
\citeauthor{xu2013study} \cite{xu2013study} and \citeauthor{booth2014automated} \cite{booth2014automated} recursively select the features based on the degree of performance degeneration when the values for the feature are permuted. 
\citeauthor{de2018advances} \cite{de2018advances} introduces several feature importance metrics for financial machine learning.
\citeauthor{article2019sun} \cite{article2019sun} maximize the mutual information between selected features and labels.
However, they do not study how to select features in conjunction with sample reweighting for better performance.

\textbf{Noise reduction for finance.}
Noise reduction is crucial to extract information from the financial data with a low signal-to-noise ratio. In this paper, we focus on denoising in the phase of model training. Apart from reweighting the samples to denoise, \citeauthor{zhang2019ieg} \cite{zhang2019ieg} and \citeauthor{xu2019l_dmi} \cite{xu2019l_dmi} design specific loss functions to denoise. Noise reduction can also be performed from the perspective of signal processing (e.g., filtering on the raw sequential data before extracting the features \cite{alrumaih2002time,al2011selecting}) or the perspective of financial risk control (e.g., controlling the extent of the risk exposure \cite{qian2006active}). 

\section{Method}
\label{sec:method}

In this section, we propose DoubleEnsemble, an ensemble model with two key components: \revise{learning} trajectory based sample reweighting and shuffling based feature selection.
We show the training process in Algorithm \ref{algorithm}.

\revise{
The training data consists of the feature matrix $X$ and the labels ${\bm y}$.
Here, $X = [{\bm x}_1, \cdots, {\bm x}_N]^T \in \mathbb{R}^{N\times F}$ is a matrix where $N$ is the number of samples, $F$ is the number of features, and ${\bm x}_i$ is the feature vector for the $i$-th sample.
${\bm y} = (y_1, \cdots, y_N)$ is a vector of size $N$ where $y_i$ is the label for the $i$-th sample.
In the process, we sequentially construct $K$ sub-models, $\mathcal{M}^{1}, \cdots, \mathcal{M}^{K}$.
After constructing the $k$-th sub-model, we define the current ensemble model $\overline{\mathcal{M}}^{k}(\cdot) = \frac{1}{k}\sum_{\kappa=1}^k \mathcal{M}^\kappa(\cdot)$ to be a simple average over the first $k$ sub-models.
The output of DoubleEnsemble is $\overline{\mathcal{M}}^{K}(\cdot)$ which is the average of all the $K$ sub-models.
}

\revise{
Each sub-model is trained based on not only the training data $(X, {\bm y})$ but also a set of selected features ${\bm f} \subseteq [F]$ and the weights ${\bm w} = (w_1, \cdots, w_N)$ where $w_i$ is the weight assigned to the $i$-th sample.
For the first sub-model, we use all the features and equal weights.
For the subsequent sub-models, we use learning trajectory based sample reweighting (SR) and shuffling based feature selection (FS) to determine the weights and select features respectively.
}

\revise{
Before we introduce the details of SR and FS, we first introduce the input for these two processes.
For SR, we retrieve the loss curves during the training of the previous sub-model and the loss values of the current ensemble.
Suppose there are $T$ iterations in the training of the previous sub-model. 
We use $C\in\mathbb{R}^{N\times T}$ to denote the loss curves where the element $c_{i,t}$ is the error on the $i$-th sample after the $t$-th iteration in the training of the previous sub-model.
For neural network,
an iteration is one training epoch, and
for boosting trees, 
we construct a new tree in an iteration.
Next, we use $L\in\mathbb{R}^{N\times 1}$ to denote the loss values where the element $l_{i}$ is the error of the current ensemble on the $i$-th sample (i.e., the error between $\overline{\mathcal{M}}^{k}({\bm x}_i)$ and $y_i$).
For FS, we directly provide the training data and the current ensemble as the input.
In the subsequent subsections we will introduce SR and FS in details.
}

\begin{algorithm}[t]
\caption{DoubleEnsemble
}
\label{algorithm}
\begin{algorithmic}[1]
\State \textbf{Input:} The training data $(X, {\bm y})$ and the number of sub-models $K$.
\State Set the weights $\revise{{\bm w}} \leftarrow (1, \cdots, 1)$
\State Select the features $\revise{{\bm f}} \leftarrow [F]$
\For{$k=1$ to $K$}
	\State $\mathcal{M}^{k} \leftarrow \text{TrainSubModel}(X, {\bm y}, {\bm w}, {\bm f})$
	\State Retrieve the \revise{loss curves $C$ and the loss values $L$}
    \State $\revise{\bm w} \leftarrow \text{SR} \left( C, L, k \right)$ \quad \quad $\triangleright$ sample reweighting
    \State $\revise{\bm f} \leftarrow \text{FS} \left( \overline{\mathcal{M}}^{k}, X, {\bm y} \right) $ \quad \quad $\triangleright$ feature selection
\EndFor
\State \textbf{Return:} The ensemble model $\overline{\mathcal{M}}^K$
\end{algorithmic}
\end{algorithm}

\textbf{Discussion.}
To extract the temporal information prior to the time point for prediction, we filter the signals (e.g., using the moving average, the Kalman filter, etc.) before calculating the features.
We empirically found that this is more effective than filtering the signals using variants of recurrent neural networks (e.g., SFM \cite{zhang2017stock}).
Besides, in our model, the prediction of the ensemble model is a simple average of the predictions from all the sub-models. 
This is a simplest yet robust way to aggregate the sub-models. 
We note it is possible to set a weight for each sub-model or develop a stacked generalization ensemble (aka stacking). 
In general, a proper way to combine the sub-models
can further improve the performance and we leave it as a future research direction.

\subsection{Learning trajectory based sample reweighting}
\label{sec:SR}

\revise{
We show the learning trajectory based sample reweighting (SR) process in Algorithm \ref{sample_reweight}. 
In the process, we first calculate the $h$-value for each sample and then divide all the samples into $B$ bins according to the $h$-value. 
Later, we assign the same weights to the samples in the same bin.
}
\begin{algorithm}[t]
\caption{SR: Learning trajectory based sample reweighting}
\label{sample_reweight}
\begin{algorithmic}[1]
\State \textbf{Inputs:} \revise{The loss curves $C$, the loss values $L$} and the index of the sub-model $k$
\State \textbf{Parameters:} The coefficients $\alpha_1$ and $\alpha_2$, the number of bins $B$ and the decay factor $\gamma$
\State Calculate the $h$-value for each sample based on \eqref{eq:reweight1}
\State Divide the samples into $B$ bins according to the $h$-values
\State Calculate the weight ${\bm w}$ based on \eqref{eq:reweight2}
\State \textbf{Return:} The weights ${\bm w}$
\end{algorithmic}
\end{algorithm}

\revise{
The calculation of the $h$-value is based on the loss curves of the previous sub-model $C$ and the loss values of the current ensemble $L$.
For robustness, we first normalize $C$ and $L$ via ranking.
The normalization function $\text{norm}:\mathbb{R}^{N\times d} \to [0,1]^{N\times d}$ replaces each element in the matrix with its rank across other elements in the column, i.e., $\text{norm}(X)_{ij} = 0.9$ if $X_{ij}$ is larger than $90\%$ of the elements in the $j$-th column of $X$.
Then, we can define normalized loss curves $\widetilde{C}=\text{norm}(C)$ and normalized loss values $\widetilde{-L}=\text{norm}(-L)$ (i.e., reversely normalized loss values).
To indicate whether the loss of a sample gets improved during the training, we compare its loss at the start and at the end of training.
We use $C_\text{start},C_\text{end} \in \mathbb{R}^{N\times 1}$ to denote the loss for all the samples at the start and at the end of training respectively.
Specifically, they are the average of the first and the last 10\% rows of $\widetilde{C}$ respectively.
For example, if we train $T=100$ iterations for each sub-model, each element in $C_\text{start}$ is the average normalized loss of a sample across the first $10$ iterations.
Next, we calculate the $h$-values for all the samples as follows:
\begin{equation}
\label{eq:reweight1}
{\bm h} = \alpha_1 
\underbrace{
\vphantom{\left(
\dfrac
{\text{norm}(C[\text{tail}])}
{\text{norm}(C[\text{head}])}
\right)}
(\widetilde{-L})}_{{\bm h}_1} 
+ \alpha_2 
\underbrace{
\text{norm} 
\left(
\dfrac
{C_\text{end}}
{C_\text{start}}
\right)}_{{\bm h}_2},
\end{equation}
where ${\bm h}, \widetilde{-L}, C_\text{start}, C_\text{end} \in \mathbb{R}^{N\times 1}$ and the operations are element-wise.
}


To avoid extreme values for the weights, we further divide the samples into $B$ bins according to the $h$-values 
and assign the same weights to the samples in the same bin. 
Suppose the $i$-th sample is divided into the $b_i$-th bin. 
The weight of this sample is assigned as follows:
\begin{equation}
\label{eq:reweight2}
w_i = \dfrac{1}{\gamma^k \langle {\bm h} \rangle_{b_i} + 0.1},
\end{equation} 
where $\langle {\bm h} \rangle_{b}$ is 
\revise{the average $h$-value for the $b$-th bin.}
Further, we use a decay factor $\gamma \in [0, 1]$ to encourage the weight distribution to be more uniform in the latter sub-models of the ensemble. 
This technique is a simplified version from the concept of the \emph{self-paced factor} in \cite{liu2019self}.

\begin{figure*}
\centering
  \includegraphics[width=0.9\linewidth]{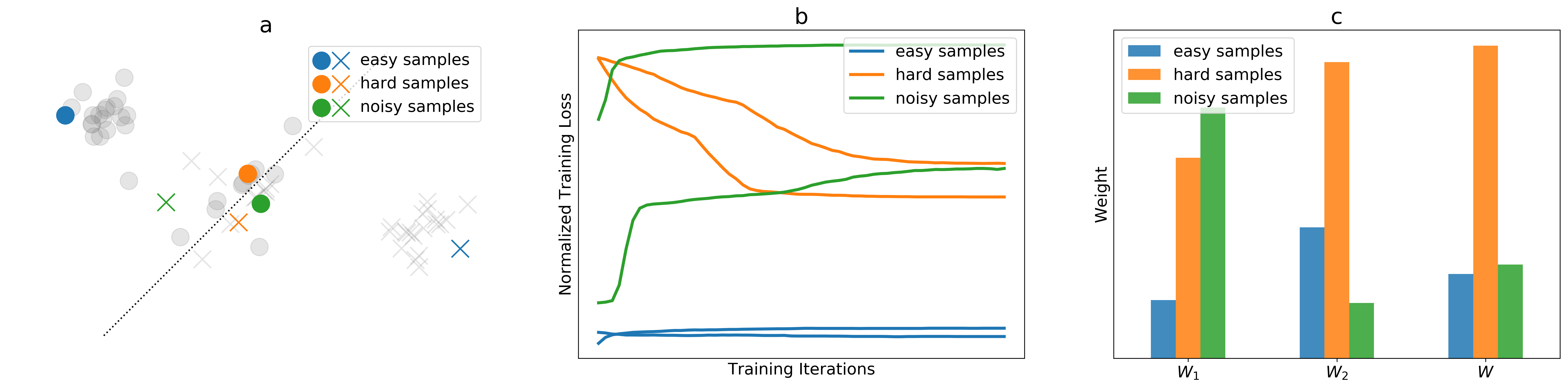}
  \caption{A toy example to illustrate the learning trajectory based sample reweighting scheme. 
  a) The example is a binary classification task with the dotted line being the underlying true decision boundary. 
  The easy samples (blue) are those with large margins to the true decision boundary while the hard samples (orange) are very close to the true decision boundary.
  There are some noisy samples (green) whose labels are random regardless of the decision boundary.
  b) 
  We train a neural network with one hidden layer as the classifier with stochastic gradient descent and plot
  the normalized 
  \revise{loss curves}
  of several samples from the three types,
  i.e., the corresponding rows in $\widetilde{C}\in [0,1]^{N\times T}$. 
  c) 
  The weights of the three types of samples calculated using Equation \eqref{eq:reweight2} with ${\bm h}_1$, ${\bm h}_2$ or ${\bm h}$ as the $h$-value, which are denoted as $W_1$, $W_2$ and $W$ respectively.
  }
  \label{fig:fig1}
\end{figure*}

Now, we explain the intuition behind our design with a
small example in Figure \ref{fig:fig1}.
Consider three types of samples in a classification task: 
the easy samples that are easily classified correctly, 
the hard samples that are close to the true decision boundary and may easily get misclassified,
and the noisy samples that may mislead the model.
We would like our reweighting scheme to boost the weights of hard samples while reducing the weights of the easy and the noisy samples, since easy samples can be fitted anyway and fitting noisy samples may lead to overfitting.
The ${\bm h}_1$ term  helps to reduce the weights of easy samples.
Specifically, the loss of an easy sample is prone to be small which leads to a large value for ${\bm h}_1$ and therefore a small weight.
However, this term also boosts the noisy samples since it is hard to distinguish the noisy samples and the hard samples solely based on the loss value. 
\revise{
Fortunately, we can distinguish them by their loss curves using ${\bm h}_2$ (cf. Figure \ref{fig:fig1}b). 
}
Intuitively, we assign large weights to the samples with a descending normalized
\revise{loss curve.}
Since the training process is driven by the majority of the samples, 
the loss of most of the samples tends to decrease while the loss of noisy samples usually keeps the same or even increases.
Therefore, the normalized 
\revise{loss curves}
of noisy samples will increase which leads to large ${\bm h}_2$ values and therefore small weights.
For easy samples, their normalized 
\revise{loss curves}
are more likely to remain the same or fluctuate slightly after a quick decay, which results in moderate ${\bm h}_2$ values and therefore moderate weights.
For hard samples, their normalized 
\revise{loss curves}
slowly decline during the training which indicates their contribution to the decision boundary.
This results in small ${\bm h}_2$ values and therefore large weights.  
We show the weights of the three types of samples calculated using ${\bm h}_1$, ${\bm h}_2$ and ${\bm h}$ as the $h$-value respectively in Figure \ref{fig:fig1}c.
We observe that, using ${\bm h}_1$ not only boosts the weights of hard samples but also those of noisy samples, while using ${\bm h}_2$ suppresses the weights of noisy samples.
With ${\bm h}_1$ and ${\bm h}_2$ combined (i.e., ${\bm h}$), we can effectively boost the hard samples and reduce the weights for the easy samples and the noisy samples.

\subsection{Shuffling based feature selection}

We use the shuffling based feature selection (FS) process in DoubleEnsemble to select features for training the next sub-model.
We show this process in Algorithm \ref{feature_select}.
\revise{
Similar to SR, we first calculate a $g$-value for each features and then divide all the features into $D$ bins according to their $g$-values. 
Later, we randomly select features from different bins with different sampling ratios.
}


\begin{algorithm}[t]
\caption{FS: shuffling based feature selection}
\label{feature_select}
\begin{algorithmic}[1]
\State \textbf{Inputs:} An ensemble model $\overline{\mathcal{M}}$, 
the training data $(X, {\bm y})$
\State \textbf{Parameters:} 
The number of bins $D$ and the sampling ratio for each bin $(r_1, \cdots, r_D)$
\State $L = \text{loss}\left(\overline{\mathcal{M}}(X), {\bm y}\right)$
\For{each feature index $f$ in $[F]$}
	\State $X_f\leftarrow$ $X$ with the $f$-th column shuffled
	\State $L_f = \text{loss}\left(\overline{\mathcal{M}}(X_f), {\bm y}\right)$
	\State $g_f =  \text{mean}\left(L_f - L \right) / \text{std}\left(L_f - L \right)$
\EndFor
\State Divide the features into $D$ bins according to the $g$-values where the $d$-th bin contains $N_d$ features
\State ${\bm f} = \emptyset $
\For{each bin $d\in [D]$}
	\State ${\bm f}_d \leftarrow \lceil r_d N_d \rceil$ features randomly sampled from the bin
	\State ${\bm f} = {\bm f} \cup {\bm f}_d$
\EndFor
\State \textbf{Return:} The selected features ${\bm f}$
\end{algorithmic}
\end{algorithm}

\revise{
$g$-value for a feature measures the contribution of this feature to the current ensemble (i.e., feature importance). 
To calculate the $g$-value for a feature, we shuffle the values of this feature and compare the losses before and after the shuffle (cf. Line 5-7 in Algorithm \ref{feature_select}).
The $g$-value for a feature is large when the elimination of 
the feature (via shuffling) significantly increases the losses on the samples, which indicates that this feature is important to the current ensemble.
For robustness against extreme $g$-values, we then divide all the features into $D$ bins according to the $g$-values 
and randomly select features from different bins with different sampling ratios (cf. Line 8-12 in Algorithm \ref{feature_select}).
The sampling ratios are preset and the ratio is large for the bin with large $g$-values.
At last, we concatenate and return all the randomly selected features. 
}

The reason for the design is as follows:
To estimate the contribution of a feature to the model, we would like to compare with the performance when the feature is absent.
One natural but costly way is to eliminate the feature, retrain and then re-evaluate the model.
Instead of training a new model, we perturb the dataset to eliminate the contribution of the feature and
compare the performance of the model using the perturbed dataset and that using the original dataset.
Our scheme computationally is more efficient since there is no need to retrain a model.

Moreover, we argue that shuffling is more appropriate than 
replacing with zeros (or the mean of the feature).
This is because many machine learning models are sensitive to the input data distribution. Shuffling keeps the marginal distribution of that feature, and replacing with zeros
completely changes the distribution.
For a simple example, consider a feature whose values are either $+1$ or $-1$ and the mean is $0$.
The trained model would focus on the regions where the feature value is $+1$ or $-1$ (regions with denser samples are better fitted). Hence, the region around feature value $0$ is not fitted well and 
the model may behave arbitrarily for samples with feature value replaced by $0$, and cannot correctly reflect the performance when this feature is eliminated.

In addition, the shuffling based feature selection method has the following advantages: 
First, it considers the contribution of the feature to the model which is trained along with other features, instead of the quality of the feature itself such as the frequently used information coefficient and information ratio \cite{goodwin1998information} in finance. 
Second, unlike other feature importance metrics that only apply to specific models (such as the information gain in boosting trees and the coefficients in Lasso \cite{santosa1986linear}), the $g$-value is applicable to different base models.

\section{Experiments}
\label{sec:experiment}

We apply DoubleEnsemble to predict for two different financial markets:
OKEx (a cryptocurrency exchange) and China's A-share market (a securities exchange).

In the first set of experiments on OKEx, we compare DoubleEnsemble with a set of baseline methods and several ablated variants of DoubleEnsemble
to measure the effectiveness of the designs in DoubleEnsemble.
Also, we design comparative experiments to quantify the robustness of our model to different level of noise.

In the second set of experiments on China's A-share market, we train predictors and then construct trading strategies based on the predictors
via variants of DoubleEnsemble and several baselines.
The experiments demonstrate that the superior performance of our predictors can be translated into the profits from the induced strategy.
We also conduct experiments under two different trading frequencies with different set of features.  

In the following experiments, we use \revise{$K=6$} sub-models. 
In the SR process, we use $\alpha_1 = \alpha_2 = 1$ and $B=10$ bins. 
In the FS process, we use $D=5$ bins and the sample ratios are $(0.8, 0.7, 0.6, 0.5, 0.4)$.

\subsection{DoubleEnsemble to trade cryptocurrencies}
\label{sec:experiment_OKEx}

\begin{table*}[ht]
    \caption{Experiment results on OKEx. See the detailed description of the experiment in Section \ref{sec:experiment_OKEx}. The numbers in each entry are the mean and the standard deviation from 5 independent runs respectively. The transaction fee is 0.2\textperthousand.}
    \label{tab:experiment1}
    \setlength{\tabcolsep}{1.6mm}{
    {\begin{tabular}{lllrrrrrrrr}
\toprule
    &                  &            & \multicolumn{4}{l}{\texttt{30\% Noise}} & \multicolumn{4}{l}{\texttt{50\% Noise}} \\
    &                  &            &     \texttt{ACC} (\%) &     \texttt{AUC} (\%) &      \texttt{F1} (\%) &      \texttt{PCT} (\textperthousand) &     \texttt{ACC} (\%) &     \texttt{AUC} (\%) &      \texttt{F1} (\%) &      \texttt{PCT} (\textperthousand) \\
\midrule
\multirow{15}{*}{MLP} & \multirow{5}{*}{DoubleEnsemble} & \texttt{SR} &  60.78/0.65 &  52.54/0.54 &  75.83/0.51 &   2.20/1.01 &  60.05/0.43 &  53.49/0.17 &  75.04/0.34 &   1.89/0.67 \\
    &                  & \texttt{SR (1st only)} &  60.93/0.17 &  52.86/0.14 &  75.72/0.13 &   2.49/0.26 &  59.95/0.44 &  52.89/0.51 &  74.96/0.34 &   1.82/0.67 \\
    &                  & \texttt{SR (2nd only)} &  60.17/1.49 &  \textbf{53.65/1.78} &  75.33/1.17 &  2.28/2.29 &  59.78/3.90 &  53.59/0.45 &  74.43/3.14 &   1.70/1.02 \\
    &                  & \texttt{FS} &  61.00/0.11 &  52.69/0.60 &  75.77/0.09 &   2.53/0.18 &  59.40/0.58 &  53.59/0.76 &  74.53/0.46 &   1.44/0.90 \\
    &                  & \texttt{SR+FS} &  \textbf{62.10/0.87} &  \textbf{53.56/0.76} &  \textbf{76.62/0.66} &  \textbf{3.18/1.35} &  \textbf{60.94/0.94} &  \textbf{54.27/0.55} &  \textbf{75.72/0.73} &  \textbf{2.49/1.44} \\
\cline{2-11}
    & \multirow{3}{*}{Basic Methods} & \texttt{SingleModel} &  58.03/0.46 &  52.57/0.39 &  73.44/0.28 &   0.50/0.60 &  58.10/0.52 &  52.68/0.51 &  74.29/0.26 &   0.73/0.52 \\
    &                  & \texttt{SimpleEnsemble} &  59.77/0.46 &  53.47/0.97 &  74.82/0.36 &  1.69/0.70 &  59.63/0.24 &  53.25/0.94 &  74.71/0.18 &   1.59/0.36 \\
    &                  & \texttt{RandomEnsemble} &  60.17/0.67 &  52.42/0.20 &  75.13/0.51 &   1.97/1.03 &  59.85/0.57 &  52.12/0.63 &  74.88/0.44 &   1.75/0.88 \\
\cline{2-11}
    & \multirow{6}{*}{Baseline Methods} & \texttt{LDMI}\cite{xu2019l_dmi} &  58.61/0.51 &  52.09/0.47 &  73.91/0.41 &   0.90/0.78 &  57.52/1.72 &  51.73/0.69 &  73.01/1.41 &  0.15/2.63 \\
    &                  & \texttt{LCCN}\cite{yao2019safeguarded}  &  57.96/0.41 &  52.80/0.53 &  73.38/0.32 &   0.45/0.62 &  58.34/0.19 &  52.27/0.47 &  73.69/0.15 &   0.71/0.30 \\
    &                  & \texttt{CoTeach}\cite{han2018co}  &  59.37/0.56 &  51.03/0.45 &  74.50/0.44 &   1.42/0.86 &  58.63/0.31 &  51.30/0.74 &  73.91/0.24 &   0.91/0.46 \\
    &                  & \texttt{MentorNet}\cite{jiang2018mentornet} &  58.37/0.40 &  52.75/0.41 &  73.71/0.32 &   0.73/0.62 &  57.92/0.41 &  52.60/0.37 &  73.35/0.33 &   0.43/0.64 \\
    &                  & \texttt{LearnReweight}\cite{ren2018learning} &  58.72/0.56 &  52.50/0.53 &  73.98/0.44 &   0.97/0.86 &  56.06/0.15 &  51.46/0.14 &  71.84/0.12 &   -0.85/0.22 \\
    &                  & \texttt{Curriculum}\cite{bengio2009curriculum} &  60.39/0.36 &  52.38/0.50 &  75.62/0.26 &   2.16/0.55 &  60.15/0.62 &  53.12/0.95 &  75.12/0.48 &   1.96/0.95 \\
\cline{2-11}
    & \multicolumn{2}{c}{No noise, \texttt{SingleModel}} & 61.20/0.82 & 52.85/0.74 & 75.93/0.62 & 2.68/1.25 & & & & \\ 
\cline{1-11}
\cline{2-11}
\multirow{10}{*}{GBM} & \multirow{5}{*}{DoubleEnsemble} & \texttt{SR} &  61.73/0.40 &  52.53/0.34 &  76.34/0.31 &   3.04/0.62 &  60.54/0.68 &  \textbf{54.33/0.28} &  75.42/0.53 &   2.22/1.05 \\
    &                  & \texttt{SR (1st only)} &  57.92/0.33 &  52.14/0.23 &  73.35/0.25 &   0.42/0.51 &  58.56/0.24 &  52.81/0.16 &  73.87/0.19 &   0.87/0.36 \\
    &                  & \texttt{SR (2nd only)} &  62.47/0.77 &  53.08/0.62 & 76.90/0.59 &  3.54/1.81 &  60.92/0.87 &  52.93/0.75 & 75.71/0.67 & 2.48/1.33 \\
    &                  & \texttt{FS} &  57.53/0.30 &  52.85/0.37 &  72.90/0.24 &   0.04/0.46 &  58.06/1.80 &  \textbf{54.40/0.58} &  73.25/1.40 &   -0.89/0.37 \\
    &                  & \texttt{SR+FS} &  \textbf{62.87/1.07} &  \textbf{54.15/0.80} &  \textbf{77.67/0.83} &  \textbf{3.83/1.64} &  \textbf{61.49/0.58} &  \textbf{53.71/0.21} &  \textbf{76.16/0.45} &   \textbf{2.87/0.90} \\
\cline{2-11}
    & \multirow{3}{*}{Basic Methods} & \texttt{SingleModel} &  56.17/0.36 &  52.71/0.46 &  71.93/0.29 &   -0.77/0.55 &  55.13/0.59 &  54.05/0.55 &  71.07/0.49 &   -1.44/1.01 \\
    &                  & \texttt{SimpleEnsemble} &  56.04/0.30 &  53.35/0.42 &  71.82/0.24 &   -0.87/0.45 &  54.42/0.19 &  54.61/0.49 &  70.48/0.16 &   -1.49/0.91 \\
    &                  & \texttt{RandomEnsemble} &  56.23/0.28 &  53.35/0.33 &  71.98/0.23 &   -0.73/0.43 &  53.62/0.28 &  54.14/0.25 &  69.81/0.23 &   -2.52/0.42 \\
\cline{2-11}
    & Baseline Methods & \texttt{Curriculum}\cite{bengio2009curriculum} &  58.31/0.58 &  52.88/0.14 &  73.67/0.46 &   1.94/0.89 &  57.24/0.29 &  53.37/0.80 &  72.34/0.23 &   0.04/0.49 \\
\cline{2-11}
    & \multicolumn{2}{c}{No noise, \texttt{SingleModel}} & 57.30/0.60 & 51.37/0.29 & 72.86/0.48 & 0.00/0.92 & & & & \\ 
\bottomrule
\end{tabular}}}
\end{table*}

This set of experiments are based on the data from OKEx.
OKEx is a cryptocurrency exchange where traders around the world can trade between different cryptocurrencies in 24 hours a day. 
In this set of experiments, we use the data from four trading pairs: ETC/BTC, ETH/BTC, GAS/BTC and LTC/BTC. 
For each trading pair, one sample corresponds to one market snapshot, which is captured for approximately every $0.3$ second.
The training samples used in the experiments are from $10$ consecutive trading days, with a total number of $3$ million. 
The testing samples come from the following $5$ trading days, with a total number of $1.5$ million.
We use $31$ features,
which are calculated based on 
the microstructure information of the market
(snapshots of the limit order book),
such as order flow imbalance (OFI) \cite{cont2014price}
and relative strength index (RSI) \cite{wilder1978new}.

We compare the algorithms under two settings with different noise levels. In the setting denoted by \texttt{30\% noise}, we add 20 additional random features and 30\% random samples (i.e., the values of these features/samples are randomly drawn from $U[0,1]$). In the setting denoted by \texttt{50\% noise}, we add 30 random features and 50\% random samples.
Next, we introduce the algorithms that we compare and the performance metrics that we use.

\textbf{DoubleEnsemble variants}

We use \texttt{SR} to denote the ensemble model that only uses the SR process, i.e., using all the features.
We use \texttt{1st only} and \texttt{2nd only} to denote the variants that only use the first term (i.e., ${\bm h}_1$) or the second term (i.e., ${\bm h}_2$) in Equation \eqref{eq:reweight1} for the SR process respectively.
We use \texttt{FS} to denote the ensemble model that only uses the FS process, i.e., using equal weights.


\textbf{Basic methods} 

\texttt{SingleModel}: 
We use the training samples with all the available features and equal weights to train a single model.
In the experiments, we use two types of base model: the neural network model (denoted as MLP) and the gradient boosting decision tree model (denoted as GBM).
For the MLP model, we use a multi-layer perceptron with 
two hidden layers (each of which has $64$ neurons) followed by a dropout layer \cite{srivastava2014dropout} and a batch-norm layer \cite{ioffe2015batch}.
We use Mish \cite{misra2019mish} as the activation function and train the model for $200$ epochs with early stopping and exponentially decaying learning rate.
For the GBM model, we use LightGBM \cite{ke2017lightgbm} with $200$ trees, each of which has at most 32 leaves.
In the later experiments, unless otherwise stated, the hyperparameters for training the sub-models are the same as used here.
Notice that this single model is the same as the first sub-model in DoubleEnsemble.

\texttt{SimpleEnsemble}:
This baseline model is an ensemble model that contains $K$ identical sub-models. The only difference between the sub-models is that they use different random seeds. 
We set this baseline to observe the performance difference brought by constructing an ensemble.

\texttt{RandomEnsemble}:
This baseline model is different from the previous baseline \texttt{SimpleEnsemble} in that, the sub-models in this baseline not only use different random seeds but also are trained with the samples assigned with random weights. 
We notice that randomly reweighting the samples may improve the performance due to the fact that it increases the diversity of the sub-models.
We set this baseline to isolate the performance different raised by the above reason.
Constructing an ensemble by randomly reweighting samples is similar to bagging where the samples are randomly selected to construct different sub-models \cite{breiman1996bagging}. 

\textbf{Baseline methods}

The following baseline methods are designed for noise robustness and we compare our algorithm with them in terms of noise sensitivity.
\texttt{LDMI} \cite{xu2019l_dmi} uses an information-theory based loss function for training a neural network robust to noisy samples.
Latent class-conditioned noise model (\texttt{LCCN})
\cite{yao2019safeguarded}
is another model designed for training a robust deep learning model against the noise by modeling the noise transition.
\texttt{CoTeaching} \cite{han2018co} simultaneously trains two neural networks and utilize the communication between the two networks to select clean data.
\texttt{MentorNet} \cite{jiang2018mentornet} trains a mentor network to weight the samples based on their training dynamics for noise reduction. 
\texttt{LearnReweight} \cite{ren2018learning} sets the weights of the samples as parameters and learns the weights via gradient descent.
The above baseline methods construct single models.
Additionally, we design \texttt{Curriculum} to construct an ensemble model with $K$ sub-models, each of which uses the $30\%$ to $100\%$ of the easiest samples (the samples with lowest losses), which can be regarded as an ensemble version of curriculum learning \cite{bengio2009curriculum}.

\textbf{Performance metrics}

\texttt{Precision}:
While standard classification problems care about the prediction accuracy on all the samples, the classification problems for financial market prediction care more about the accuracy for the retrieved samples. 
In financial market prediction, a retrieved sample corresponds to a trading signal and therefore relates to the profit of the trading strategy.
Hence, we set the threshold such that approximately 1\% of the samples are retrieved, and use precision as the performance metric. 
This corresponds to trading each pair for every $30$ seconds on average.

\texttt{AUC}:
We also use the area under the ROC curve (ROC AUC) as the performance metric to summarize the performances of the predictor under different thresholds.

\texttt{F1}:
In financial market prediction, we also care about the recall, which indicates the ability of the model to seize the trading opportunity. Therefore, we also use the F1 score as the performance measure which integrates the precision and the recall and it is defined as
$F1 = 2/(\text{precision}^{-1} + \text{recall}^{-1})$.

\texttt{PCT}:
Finally, we directly measure the profitability by \texttt{PCT}, which is the average return for each trading day if we follow the following strategy.
Each time the sample corresponding to the current trading time point is retrieved by the predictor (which we call a trading signal), we long the base currency in the next trading time point and then close the position after $20$ seconds.

\begin{figure*}[hbtp]
    \centering
    
    \begin{subfigure}[b]{0.43\textwidth}
        \centering
        \includegraphics[width=\textwidth]{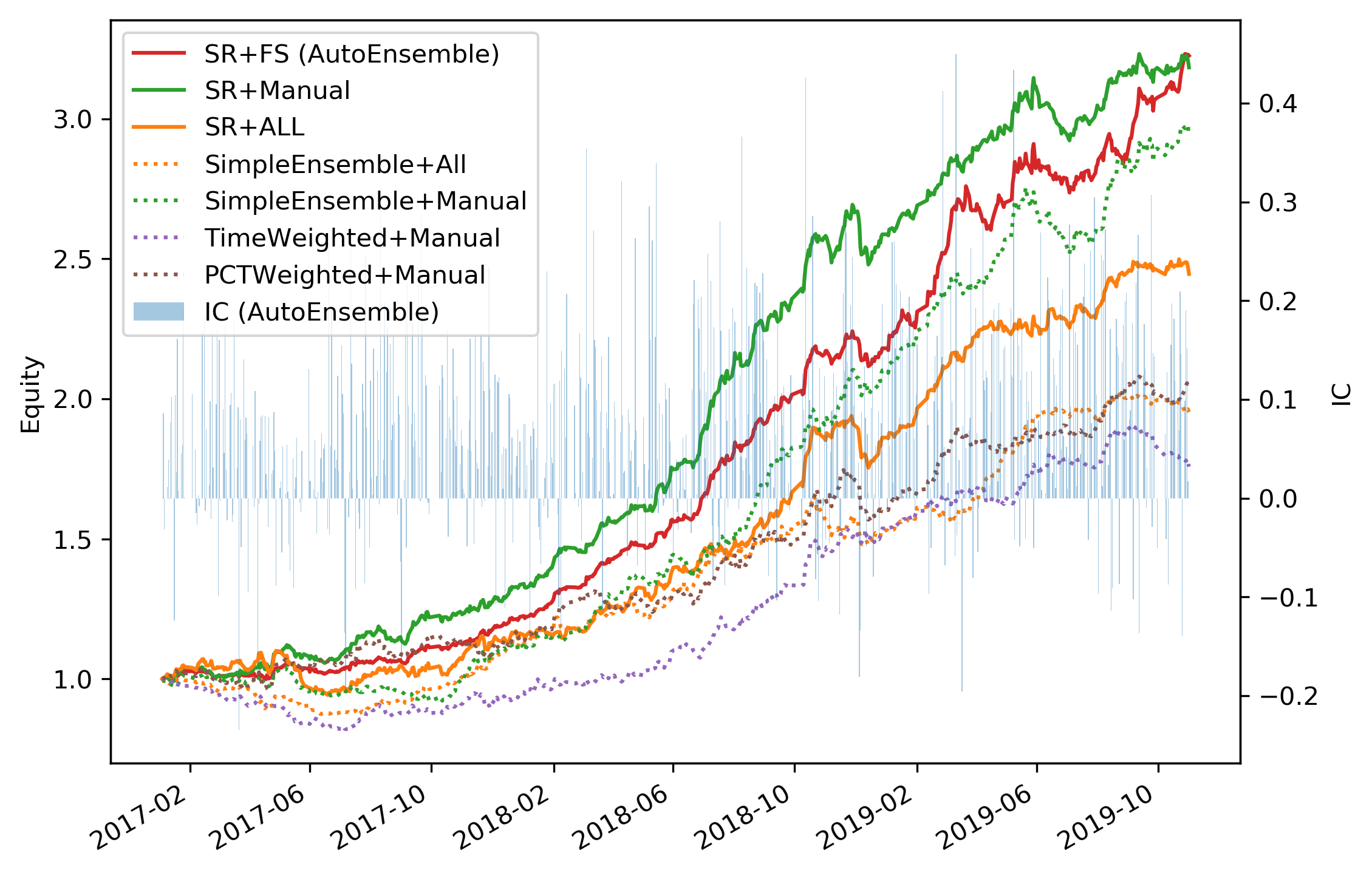}
        \caption{Results in the \texttt{DAILY} setting using MLP as the base model.}
        \label{fig:figure1}
    \end{subfigure}
    \hspace{10pt}
    \begin{subfigure}[b]{0.43\textwidth}
        \centering
        \includegraphics[width=\textwidth]{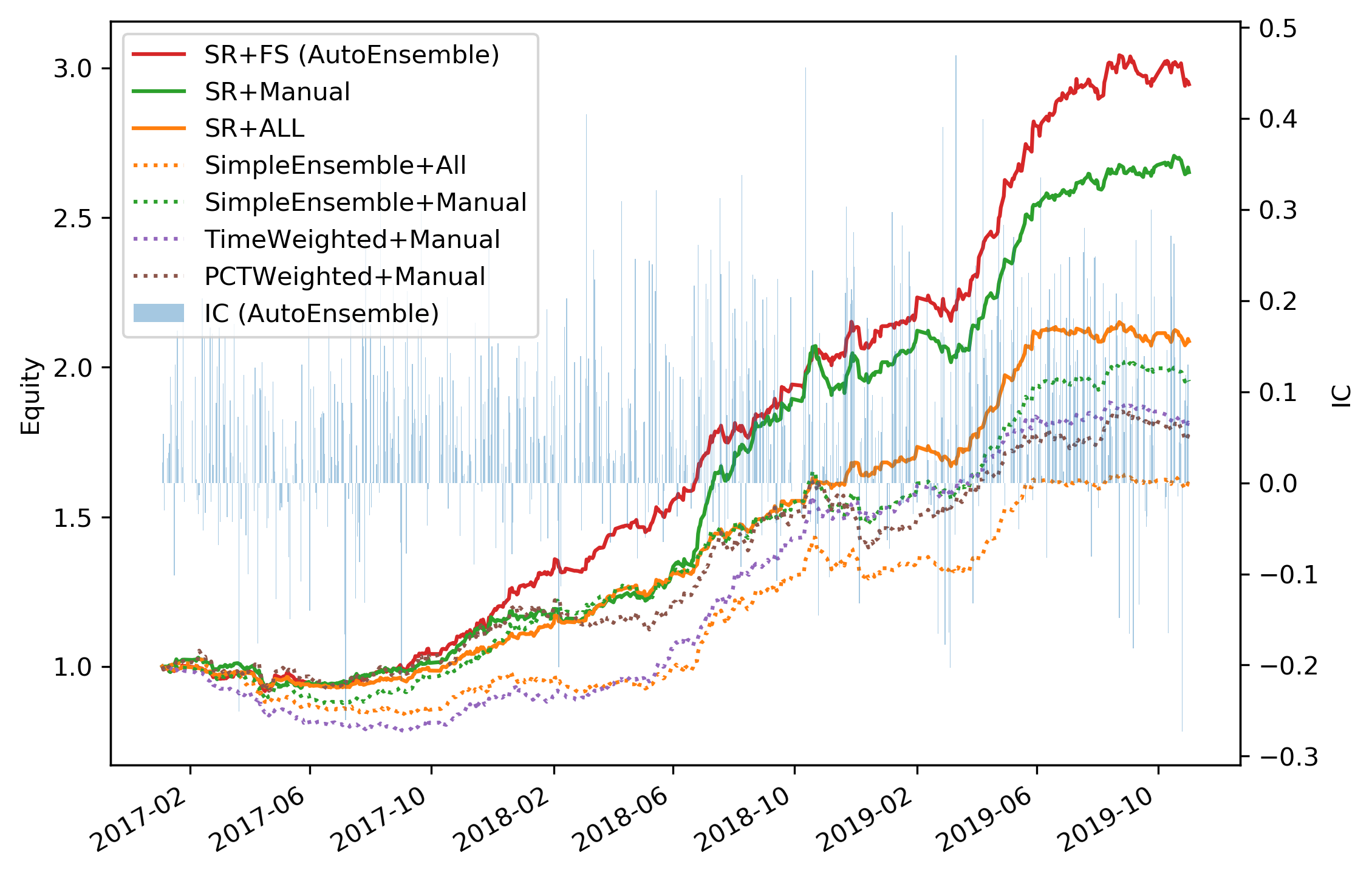}
        \caption{Results in the \texttt{DAILY} setting using GBM as the base model.}
        \label{fig:figure2}
    \end{subfigure}
    \begin{subfigure}[b]{0.43\textwidth}
        \centering
        \includegraphics[width=\textwidth]{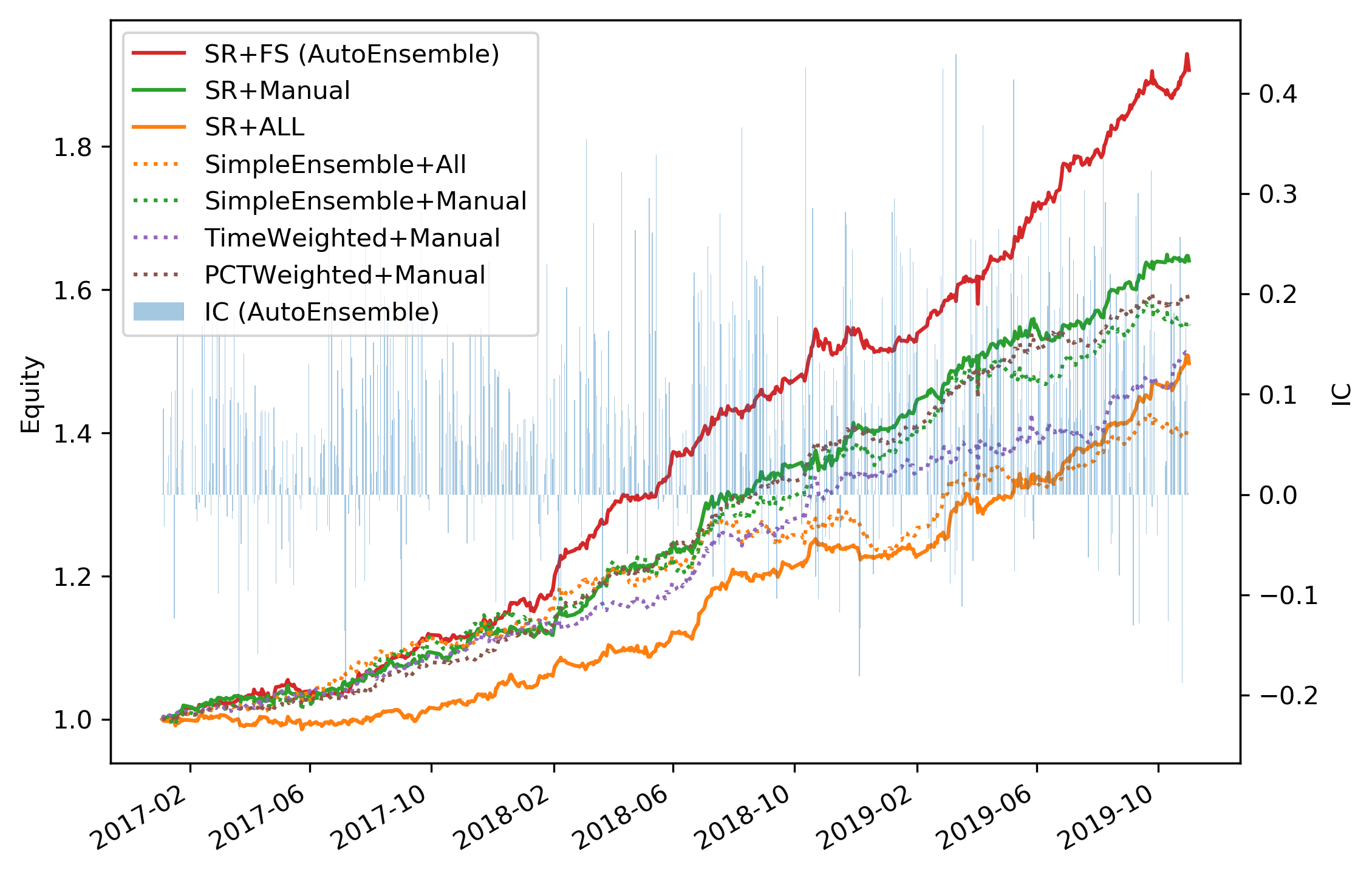}
        \caption{Results in the \texttt{WEEKLY} setting using MLP as the base model.}
        \label{fig:figure3}
    \end{subfigure}
    \hspace{10pt}
    \begin{subfigure}[b]{0.43\textwidth}
        \centering
        \includegraphics[width=\textwidth]{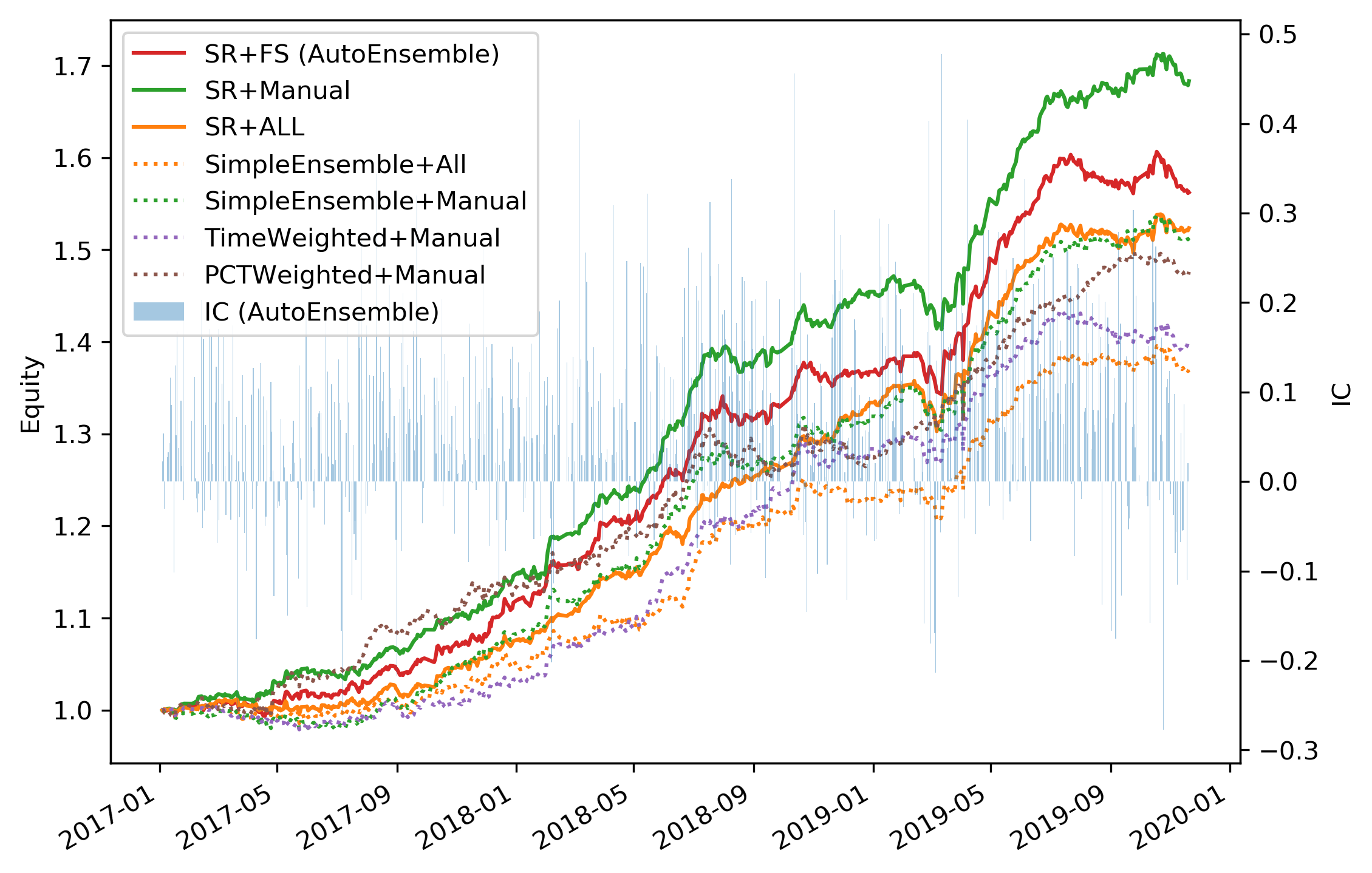}
        \caption{Results in the \texttt{WEEKLY} setting using GBM as the base model.}
        \label{fig:figure4}
    \end{subfigure}
    
    \caption{Hedged equity curves of different models under different settings. The transaction fee for each pair of trading is $0.3\%$. The blue bars in the background indicate the ICs for the overall prediction. 
    }
    \label{fig:experiment2}
\end{figure*}

\textbf{Experiment results}

We show the experiment results for the cryptocurrency prediction in Table \ref{tab:experiment1}. 
The first number in each entry is the mean of 5 runs with different random seeds, and the second number in the entry is the standard deviation of the 5 runs.

First, we observe that the DoubleEnsemble variants achieve a good performance in the two settings with different noise levels and the DoubleEnsemble algorithm (i.e., \texttt{SR+FS}) achieves the best performance.
Besides, although the AUC difference between the DoubleEnsemble variants and other baselines is not significant, the precision and the profitability difference is notable. 
This indicates that DoubleEnsemble has a higher accuracy on the key samples (i.e., the distinguishable samples with high future returns) and therefore is more suitable for financial applications.

Second, the experiment result also demonstrates the role of the SR process. 
We can compare the SR models (the models that use SR) with \texttt{SingleModel}, \texttt{SimpleEnsemble} and \texttt{RandomEnsemble}. 
When using MLP as the base model, the performance improvement brought by the SR process not only comes from constructing an ensemble or the diversity increase resulted from reweighting, but also comes from the reweighting scheme used in the SR process. 
When using GBM as the base model, the performance improvement is mainly resulted from the reweighting scheme of the SR process. 
This quantifies the important role that SR plays in identifying and weighting the key samples.
We also found that, although some baselines (such as \texttt{LCCN}) are robust to different noise levels, the SR models outperform the previous baseline methods that reweight the samples to denoise.
The reason may be that the SR process is designed not only to denoise but also to promote the performance by boosting the key samples. 

At last, the experiment result shows the performance improvement brought by the FS process. 
We can observe the improvement brought by the FS process by comparing \texttt{FS} with \texttt{RandomEnsemble} or by comparing \texttt{SR+FS} with \texttt{SR}. 

\subsection{DoubleEnsemble to trade stocks}

\begin{table*}[ht]
    \caption{Performance of the stock trading strategies. The transaction fee is 0.3\%.}
    \label{tab:experiment2}
    \setlength{\tabcolsep}{1.5mm}{
    {\begin{tabular}{llllrlllrll}
\toprule
    &                 &                    & \multicolumn{4}{l}{\texttt{DAILY}} & \multicolumn{4}{l}{\texttt{WEEKLY}} \\
    &                 &                    & \texttt{Ann.Ret.} & \texttt{Sharpe} &    \texttt{MDD} &        \texttt{IC/IR} & \texttt{Ann.Ret.} & \texttt{Sharpe} &    \texttt{MDD} &        \texttt{IC/IR} \\
\midrule
\multirow{7}{*}{MLP} & \multirow{3}{*}{DoubleEnsemble} & \texttt{SR+FS} &    \textbf{51.37\%} &  \textbf{4.941} &  \textbf{5.98\%} &  \textbf{0.115/1.035} &    \textbf{25.67\%} &  \textbf{4.448} &  \textbf{2.41\%} &  \textbf{0.078/0.773} \\
    &                 & \texttt{SR+Manual} &    50.68\% &  4.343 &  7.94\% &  0.106/0.994 &    19.16\% &  3.300 &  2.48\% &  0.078/0.784 \\
    &                 & \texttt{SR+ALL} &    37.25\% &  2.933 &  14.34\% &  0.103/0.966 &    15.36\% &  3.051 &  2.32\% &  0.070/0.691 \\
\cline{2-11}
    & \multirow{4}{*}{Baselines} & \texttt{SimpleEnsemble+All} &    26.74\% &  2.435 &  12.61\% &  0.091/0.967 &    12.56\% &  2.049 &  4.59\% &  0.058/0.670 \\
    &                 & \texttt{SimpleEnsemble+Manual} &    46.49\% &  3.813 &  11.75\% &  0.097/0.963 &    16.78\% &  2.817 &  2.45\% &  0.068/0.757 \\
    &                 & \texttt{TimeWeighted+Manual} &    22.10\% &  1.936 &  18.49\% &  0.081/0.791 &    15.10\% &  2.342 &  3.56\% &  0.061/0.700 \\
    &                 & \texttt{PCTWeighted+Manual} &    28.65\% &  2.269 &  10.32\% &  0.094/0.940 &    17.07\% &  3.704 &  2.84\% &  0.070/0.735 \\
\cline{1-11}
\cline{2-11}
\multirow{7}{*}{GBM} & \multirow{3}{*}{DoubleEnsemble} & \texttt{SR+FS} &    \textbf{46.60\%} &  \textbf{4.151} &  \textbf{8.60\%} &  \textbf{0.103/0.861} &    16.77\% &  3.160 &  3.23\% &  0.068/0.668 \\
    &                 & \texttt{SR+Manual} &    41.24\% &  3.854 &  9.87\% &  0.096/0.807 &    \textbf{19.84\%} &  \textbf{3.862} &  \textbf{3.93\%} &  \textbf{0.071/0.676} \\
    &                 & \texttt{SR+ALL} &    29.75\% &  3.594 &  7.13\% &  0.097/0.816 &    15.76\% &  3.379 &  4.04\% &  0.070/0.670 \\
\cline{2-11}
    & \multirow{4}{*}{Baselines} & \texttt{SimpleEnsemble+All} &    18.19\% &  1.661 &  18.45\% &  0.101/0.858 &    11.55\% &  2.337 &  3.61\% &  0.065/0.635 \\
    &                 & \texttt{SimpleEnsemble+Manual} &    26.74\% &  2.435 &  12.61\% &  0.097/0.815 &    15.48\% &  2.902 &  3.52\% &  0.068/0.650 \\
    &                 & \texttt{TimeWeighted+Manual} &    23.39\% &  2.176 &  21.72\% &  0.093/0.768 &    12.47\% &  2.498 &  3.13\% &  0.062/0.636 \\
    &                 & \texttt{PCTWeighted+Manual} &    22.20\% &  1.669 &  13.68\% &  0.093/0.832 &    14.49\% &  2.355 &  4.22\% &  0.066/0.642 \\
\bottomrule
\end{tabular}
}}
\end{table*}

In this set of experiments, we train predictors for the stock market and trade the stocks based on the prediction.
We base our experiments on China's A share market where over 3,000 stocks are traded.
Each sample corresponds to one trading day of one stock.

\textbf{Experiment settings}

We conduct experiments in two different settings.
In the first setting (denoted by \texttt{DAILY}), we long the top 20 stocks suggested by the predictor at the market closing of each trading day, and then sell these stocks upon the closing time of the next trading day. 
The predictions are based on 182 features that are calculated 3 minutes before the market closing of that trading day. 
In the second setting (denoted by \texttt{WEEKLY}), after the market closing on each trading day, we calculate 254 features based on the historical market information and make the prediction.
In the next trading day, we long the top 10 stocks suggested by the prediction at the open price and hold these stocks for five trading days.
Thereafter, we sell these stocks after the opening of the fifth trading day.
In this setting, we are holding 50 stocks for most of the time.
The features in 
both settings
are composed of technical factors and fundamental factors, such as moving average convergence/divergence (MACD) \cite{appel2005technical} and price-to-book ratio (P/B) \cite{chan1991fundamentals}.
They are designed for the prediction at different frequencies and created by different trading firms. 
Therefore, they possess quite different underlying properties.
Since there are more features in this experiment, we use three hidden layers with more neurons (256, 128 and 64 neurons respectively) in the MLP model and 250 trees in the GBM model.

We run the backtests for the models following a \emph{rolling} scheme described as follows.
We re-train the model every week and use the features of the latest 500 trading days (i.e., approximately the latest two years) each time we train the model. The trading period for two settings is from 
January 2017 to November 2019. 
{
For trading details, 
we exclude the stocks that reach daily surged limit or listed within 3 months.
We long the top $N$ stocks with equal weights.
The transaction fee plus slippage is 0.3\%.
We did not particularly consider the impact of holidays and suspension when making predictions and conducting backtest.
}

\textbf{Models}

In this set of experiments, we compare the DoubleEnsemble variants with a set of baselines.

In terms of sample reweighting, we compare the SR process with \texttt{SimpleEnsemble} and two other heuristic reweighting schemes designed for financial market prediction.
Based on the observation that the patterns in the market varies with time, \texttt{TimeWeighted} gives larger weights to more recent samples to encourage the model to exploit current patterns.
Also, since we care about the accuracy on the samples that trigger trading signals, the model should pay more attention to the samples that are possibly retrieved.
Accordingly, we design and compare to \texttt{PCTWeighted} where the historical samples with high returns are assigned with larger weights.
{We use \texttt{PCT} to refer to the percentage of price movement, i.e., return.
}.

In terms of feature selection, we compare the FS process with the baseline that uses fixed manually selected features (\texttt{Manual}) or uses all features without selection (\texttt{All}). 
The manually selected features are obtained based on a careful analysis on various aspects of the features, such as the historical performance, the information source and the risk.
The two set of features (for \texttt{DAILY} and \texttt{WEEKLY} respectively) are used in the real trading and shown to be stable and effective in the real practice.

{
\textbf{Performance metrics}
}

\texttt{Ann.Ret.}:
We use the hedged annualized return to measure how much return the investment portfolio constructed by the model earned exceeds the market.
We divide our daily funds into two equal parts to buy stocks and hedge the market respectively. 
To hedge the market, 
we short the corresponding stock index futures.
Moreover, we consider the compound return, i.e.,
$(1+\texttt{Ann.Ret.})^n = \texttt{Total.Ret.}$ where \texttt{Total.Ret.} the return during $n$ years.

\texttt{Sharpe}: 
The Sharpe ratio is one of the most commonly used metrics for stock investment, it reflects the risk adjusted profitability.
Specificaly, $\texttt{Sharpe}=\texttt{Ann.Ret.}/\texttt{Ann.Vol.}$, where \texttt{Ann.Vol.} is annualized volatility.

\texttt{MDD}: 
Maximum drawdown (MDD) is the maximum relative loss from a peak to a trough for a portfolio.
MDD is an indicator of downside risk over a specified time period. 
MDD is related to investors' maximum affordability and needs to be kept as low as possible.

\texttt{IC/IR}: 
The information coefficient (IC) and information ratio (IR) indicate the quality of the prediction.
In our experiments, we use $\text{IC}_{daily} = \text{corr}(\text{rank}(Y_{pred})/\text{rank}(Y_{true}))$ and $\text{IC}=\text{mean}(\text{IC}_{daily})$, $\text{IR}=\text{IC}/\text{std}(\text{IC}_{daily})$, where $Y_{pred}$ is the prediction and $Y_{true}$ is the truth, $\text{IC}_{daily}$ is the IC for each time step.

\textbf{Experiment results}

We run backtests for the aforementioned models and hedge the systemic risk of the market by holding a short position of the corresponding exchange traded funds (ETF).
We plot the hedged equity curves for these models under different settings in Figure \ref{fig:experiment2}.
We also list the performance measures of the the backtest results in Table \ref{tab:experiment2}. 

In 
Figure \ref{fig:experiment2}, we show four sets of experiments. 
The four sets of experiments are conducted under different settings (\texttt{DAILY} or \texttt{WEEKLY}) and using different base models (MLP or GBM).
The curves in the figure are the hedged equity curves for different models, and the blue bars in the background indicate the information coefficient (\texttt{IC}) of the \texttt{SR+FS} model on each trading day. 
The information coefficient on a trading day is the Spearman's rank correlation coefficient between the continuous signals outputted by the model on that trading day and the actual future returns. 
While the equity curve reflects the prediction accuracy on the top retrieved samples, the information coefficient reflects the prediction accuracy on all the samples

We can see that the performance of \texttt{SR+FS} (the red lines) is better than that of \texttt{SR+ALL} (the orange lines) where all the features are used in each of the sub-models without selection. 
This indicates the effectiveness of the FS process. 
However, the automatic feature selection by the FS process is not as good as the manually selected features, which is quite a strong benchmark. 
We leave it as a future research direction to discover an automatic end-to-end feature selection method that is comparable or better than the manual selection.

Moreover, we observe that the models with the SR process achieve better performances than the models without the SR process (i.e., \texttt{SimpleEnsemble}). 
This can be observed by comparing the \texttt{SR+Manual} model (green solid line) with the \texttt{SimpleEnsemble+Manual} model (green dashed line) or by comparing the \texttt{SR+ALL} model (orange solid line) with the \texttt{SimpleEnsemble+ALL} (orange dashed line).
This indicates that the SR process can improve the performance by paying more attention to the key samples.

At last, we observe that the performance of \texttt{PCTWeighted} and \texttt{TimeWeighted} is even not as good as that of \texttt{SimpleEnsemble} in most of the settings, 
except that \texttt{PCTWeighted+Manual} is better than \texttt{SimpleEnsemble+Manual} in the \texttt{WEEKLY} setting when using MLP as the base model.
Also, the performance of these two reweighting schemes varies largely across different settings or different base models. 
The effectiveness of paying attention to the near samples or the samples with high future returns depends on the market environment. 
For example, if the market environment changes quickly, paying attention to the near samples may avoid the interference of the past samples which represent different market patterns. 
Paying attention to the samples with high future returns corresponds to the emphasis on the positive samples instead of all the samples. 
This may improve the precision when the market environment is stable.
Compared with these two heuristic reweighting schemes, the SR process weights the samples in a self-paced style and therefore is more robust across different settings.

In Table \ref{tab:experiment2}, we use the hedged annualized return (\texttt{Ann.Ret.}), the Sharpe ratio (\texttt{Sharpe}), the maximum drawdown (\texttt{MDD}), the mean of the ICs (\texttt{IC}) and the information ratio (\texttt{IR}) as the performance measure for the trading strategies.
The information ratio is the mean of the ICs divided by the standard variation of the ICs.

We found that DoubleEnsemble (\texttt{SR+FS}) achieves an annualized return of more than 50\% with low risk. 
The Sharpe ratio is near $5.0$ and the maximum drawdown is less than $6.0\%$. 
This demonstrate that the strategy induced by DoubleEnsemble has a superior and stable performance.

{
\subsection{Discussion on computational complexity}
}

\revise{
First, we observe that sample reweigting (SR) and feature selection (FS) do not significantly increase the training time.
Compared with the training sub-models, the cost of the SR and FS process is negligible. 
Indeed, each FS process uses the existing model to predict multiple times. 
However, it does not involve additional training and the prediction time is generally far less than the training time of a sub-model.
Second, in financial applications, the model can be trained offline and is embedded in real-time trading systems where latency may lead to slippage.
Therefore, 
we care more about the execution time instead of the training time.
In terms of the execution time of the whole process, we find the main constraint is the calculation of factors instead of the model prediction in practice. 
Moreover, the sub-models in the ensemble can predict in parallel to avoid the additional time cost induced by using an ensemble model. 
}


\section{Conclusion}
\label{sec:conclusion}

In this paper, we proposed a robust and effective ensemble model, DoubleEnsemble, via learning trajectory based sample reweighting and shuffling based feature selection for financial market prediction.
The \emph{learning trajectory based sample reweighting}
assigns the samples of different difficulty with different weights,
and hence is particularly suitable for highly noisy and irregular market data.
The \emph{shuffling based feature selection} can identify the contribution of the features to the model and select important and divers features for different sub-models.
We conducted experiments on two different financial markets and
compared DoubleEnsemble with several ablated variants and baseline methods. 
Our experiments demonstrate that the designs in DoubleEnsemble are effective and lead to a profitable and robust trading strategy. 

\section*{Acknowledgement}

Jian Li and Chuheng Zhang are supported in part by the National Natural Science Foundation of China Grant 61822203, 61772297, 61632016, 61761146003, and the Zhongguancun Haihua Institute for Frontier Information Technology, Turing AI Institute of Nanjing and Xi'an Institute for interdisciplinary information core Technology.
{
Yuanqi Li is supported by National Key R$\&$D Program of China No.2017YFC082070 from E-hualu.
}
Xi Chen is supported by NSF via Grant IIS-1845444.

\balance
\bibliographystyle{IEEEtranN}
\bibliography{sample-base}

\end{document}